\documentclass[runningheads]{llncs}
\usepackage{cite}
\usepackage{amsmath,amssymb,amsfonts}
\usepackage{textcomp}
\usepackage{booktabs}
\usepackage{graphicx}
\usepackage{xcolor}
\usepackage{enumitem}
\usepackage{rotating}
\usepackage{multirow}
\usepackage{array}
\usepackage{hyperref}

\begin{document}
\title{Efficient Learning of Fuzzy Logic Systems for Large-Scale Data Using Deep Learning
\thanks{A. Köklü and Y. Güven contributed equally to this work. This work was supported by MathWorks\textsuperscript{\textregistered} in part by a Research Grant awarded to T. Kumbasar.}}

\titlerunning{Efficient Learning of FLSs for Large-Scale Data Using DL}

\author{Ata Köklü\orcidID{0009-0007-3729-0514} \and
Yusuf Güven\orcidID{0009-0009-4303-6445}  \and
Tufan Kumbasar\orcidID{0000-0001-9366-0240}}

\authorrunning{Ata Köklü et al.}

\institute{Istanbul Technical University, Istanbul 34437, Turkey \\
\email{\{koklu18, guveny18, kumbasart\}@itu.edu.tr}}

\maketitle  
\begin{abstract}

Type-1 and Interval Type-2 (IT2) Fuzzy Logic Systems (FLS) excel in handling uncertainty alongside their parsimonious rule-based structure. Yet, in learning large-scale data challenges arise, such as the curse of dimensionality and training complexity of FLSs. The complexity is due mainly to the constraints to be satisfied as the learnable parameters define FSs and the complexity of the center of the sets calculation method, especially of IT2-FLSs. This paper explicitly focuses on the learning problem of FLSs and presents a computationally efficient learning method embedded within the realm of Deep Learning (DL). The proposed method tackles the learning challenges of FLSs by presenting computationally efficient implementations of FLSs, thereby minimizing training time while leveraging mini-batched DL optimizers and automatic differentiation provided within the DL frameworks. We illustrate the efficiency of the DL framework for FLSs on benchmark datasets.

\keywords{Fuzzy logic systems,  deep learning, learning,  big data}
\end{abstract}

\section{Introduction}

Type-1 (T1) and Interval Type-2 (IT2) Fuzzy Logic Systems (FLSs) are widely used in various tasks \cite{mendel_kitap, pekaslan2019leveraging}. In particular, it has been shown that IT2-FLSs are powerful tools for representing uncertain and nonlinearity since they employ and process IT2 Fuzzy Sets (FSs) as Membership functions (MFs) \cite{wiktorowicz2023t2rfis,sakalli}. The ability of IT2-FLSs to manage uncertainty hinges on their inference and structure, which are pivotal components determining how uncertainty is dealt with \cite{mendel_kitap, beke}. Uncertainty processing primarily relies on the center of sets calculation methods \cite{beke}. Karnik-Mendel Algorithm (KMA) is one of the most used methods, yet it comes with inference complexity and requires high computational time due to the iterative and sorting process\cite{complex}. This results in complexity in learning large-scale data in addition to the curse of dimensionality problem \cite{HTSK, xue2023high, Price}. Yet, the learning methods for IT2-FLSs are just extensions of the ones of T1-FLSs \cite{wiktorowicz2023t2rfis, shihabudheen2018recent}. Recently, to tackle training challenges, DL optimizers and structures have been fused with FLSs \cite{Deep_fuzzy, Price, beke, perez}.

This paper presents a computationally efficient learning method embedded within the realm of Deep Learning (DL). We first provide parameterization tricks for the Learnable Parameters (LPs) so that training via unconstrained DL optimizers is feasible. Then, we focus on the inference computation of FLSs and present efficient mini-batched inferences. We eliminate the iterative nature of KMA by presenting a parallel computing approach to process mini-batches efficiently. Thus, the proposed method minimizes training time while leveraging optimizers and automatic differentiation provided within the DL frameworks. We illustrate the efficiency of the DL framework for FLSs on benchmark datasets.

The paper is organized as follows. Section II provides background on FLSs. Section III presents the DL framework for FLSs. Section IV provides the comparative analysis while Section V gives the driven conclusions and future work. 

\section{Background on T1 and IT2 FLSs}
The rule structure of a T1-FLS composed of $P$ rules $(p=1,2, \ldots, P)$ and defined with an input $\boldsymbol{x}=\left(x_{1}, x_{2}, \ldots, x_{M}\right)^{T}$  and an output $y$ is as follows:  

\begin{equation}\label{rule_T1}
R_{p}: \text{If } x_{1} \text{ is } {A}_{p, 1} \text{ and} \ldots x_{M} \text{ is } {A}_{p, M} \text{ Then } y \text{ is } y_p
\end{equation}
while the one of an IT2-FLS processing IT2-FSs ($\tilde{A}_{p, m}$) is as:
\begin{equation}\label{rule_IT2}
R_{p}: \text{If } x_{1} \text{ is } \tilde{A}_{p, 1} \text{ and} \ldots x_{M} \text{ is } \tilde{A}_{p, M} \text{ Then } y \text{ is } y_p
\end{equation}
where $y_p$ is $y_{p}=\sum\nolimits_{m=1}^{M} a_{p, m} x_{m}+a_{p, 0}$.

The output calculation of the T1-FLS as follows:

\begin{equation}\label{y_T1}
y_{T 1}(\boldsymbol{x})=\frac{\sum_{p=1}^P f_p(\boldsymbol{x}) y_p }{\sum_{p=1}^P f_p(\boldsymbol{x})}
\end{equation}
where ${f}_{p} = \prod_{m=1}^{M}{\mu}_{A_{p,m}}$ denotes the rule firing of the $p^{th}$. Here, ${\mu}_{A_{p,m}}$ is the antecedent MF and defined with Gaussian T1-FS as:
\begin{equation}\label{T1_antecedent}
{\mu}_{{A}_{p, m}}\left(x_m\right) =  \exp \left(-\left(x_m-{c}_{p, m}\right)^2 / 2 {\sigma}_{p, m}^2\right)
\end{equation}
Here, $c_{p,m}$ is the center while $\sigma_{p, m}$ is the standard deviation of MFs. 

The output of IT2-FLS is defined as:
\begin{equation}
    y_{IT2}(\boldsymbol{x}) = (\underline{y}_{IT2}(\boldsymbol{x}) + \overline{y}_{IT2}(\boldsymbol{x})) / 2
\end{equation}
and $\underline{y}_{IT2}$ and $\overline{y}_{IT2}$, bounds of the type reduced set, are obtained by optimizing:
\begin{equation}\label{IT2_lower_upper}
\begin{split} 
\underline{y}_{IT2}(\boldsymbol{x}) = \min _{L \in [1, P-1]}{\frac{\sum_{p=1}^{L} \overline{f}_{p}(\boldsymbol{x}) {y}_{{p}} + \sum_{p=L+1}^{P} \underline{f}_{p}(\boldsymbol{x}) {y}_{{p}}}{\sum_{p=1}^{L} \overline{f}_{p}(\boldsymbol{x}) + \sum_{p=L+1}^{P} \underline{f}_{p}(\boldsymbol{x})} }\\
\overline{y}_{IT2}(\boldsymbol{x}) = \max _{R \in [1, P-1]}\frac{\sum_{p=1}^{R} \underline{f}_{p}(\boldsymbol{x}) {y}_{{p}} + \sum_{p=R+1}^{P} \overline{f}_{p}(\boldsymbol{x}) {y}_{{p}}}{\sum_{p=1}^{R} \underline{f}_{p}(\boldsymbol{x}) + \sum_{p=R+1}^{P} \overline{f}_{p}(\boldsymbol{x})}
\end{split}
\end{equation}
where $L, R$ are the switching points of the KMA. $\underline{f}_{p}$ and $\overline{f}_{p}$ are the lower and upper rule firing of the $p^{th}$ rule and are defined as:

\begin{equation}\label{IT2_firing}
    \underline{f}_{p} = \prod_{m=1}^{M}\underline{\mu}_{\tilde{A}_{p,m}}\text{ , }\overline{f}_{p} = \prod_{m=1}^{M}\overline{\mu}_{\tilde{A}_{p,m}}
\end{equation}
Here, $\overline{\mu}_{\tilde{A}_{p,m}}$ and $\underline{\mu}_{\tilde{A}_{p,m}}$ are the UMFs and LMFs, respectively:
\begin{equation}\label{T1_antecedent}
\begin{split}
    \overline{{\mu}}_{\tilde{A}_{p, m}}\left(x_m\right) =  \exp \left(-\left(x_m-{c}_{p, m}\right)^2 / 2 {\overline{\sigma}}_{p, m}^2\right)\\
    \underline{{\mu}}_{\tilde{A}_{p, m}}\left(x_m\right) =  h_{p, m}\exp \left(-\left(x_m-{c}_{p, m}\right)^2 / 2 {\underline{\sigma}}_{p, m}^2\right)
\end{split}
\end{equation}
where $c_{p, m}$ is the center, $\tilde{\sigma}_{p, m}=[\underline{\sigma}_{p, m},\overline{\sigma}_{p, m}]$ is the standard deviation while $h_{p, m}$ defines the height of the LMF $\forall p, m$.

\section{Training FLSs within DL Frameworks} 

To train the FLS with mini-batched DL optimizers, we partition the dataset $S = \left\{\boldsymbol{x}_{n}, y_{n}\right \}_{n=1}^{N}$, where $\boldsymbol{x}_n=\left(x_{n, 1}, \ldots, x_{n, M}\right)^{T}$, and $y_{n}$, into $K$ mini-batches containing $B$ samples as $S=\left\{\left(\boldsymbol{x}_{1:B}^{(1)}, {y}_{1:B}^{(1)}\right), \ldots,\left(\boldsymbol{x}_{1:B}^{(K)}, {y}_{1:B}^{(K)}\right)\right\}$.

At each epoch, the following optimization problem is minimized via a DL optimizer:
\begin{equation} \label{loss_cons}
    \min _{\boldsymbol{\theta} \in \mathcal{C}} L=\frac{1}{B}\sum_{n=1}^{B}\left(L_{R}\left(x_n, y_n\right)\right)
\end{equation}
where $C$ represents the constraint set of LPs and $L_R(\cdot)$ is the empirical loss function, e.g. L1, L2, ...etc.  

In the remaining part of the section, we present the core components of the DL framework to learn FLSs, which are:
\begin{itemize}
    \item Eliminating $\boldsymbol{\theta} \in \mathcal{C}$ for seamless deployment of unconstrained DL optimizers. 
    \item Developing efficient mini-batch FLS inferences for training. 
\end{itemize}

\subsection{Handling the Constraints}

The LP set of the FLSs comprise two subsets: $\boldsymbol{\theta}_{\boldsymbol{A}}$ for antecedent MFs and $\boldsymbol{\theta}_{\boldsymbol{C}}$ for consequent MFs. For both FLSs, $\boldsymbol{\theta}_{\boldsymbol{C}}$ is defined as $\boldsymbol{\theta}_{\mathrm{C}} = \{ \boldsymbol{a}, \boldsymbol{a}_{0} \}$, where $\boldsymbol{a} \in \mathbb{R}^{P \times M}$ and $\boldsymbol{a}_{0} \in \mathbb{R}^{P \times 1}$, while $\boldsymbol{\theta}_{\boldsymbol{A}}$ is defined for  

\begin{itemize}
    \item T1-FLS as $\boldsymbol{\theta}_{\boldsymbol{A}} =\{\boldsymbol{c}, \boldsymbol{\sigma}\}$ where $\boldsymbol{c} \in \mathbb{R}^{P \times M}$, $\boldsymbol{\sigma} \in \mathbb{R}^{P \times M}$.
    \item IT2-FLS as $\boldsymbol{\theta}_{\boldsymbol{A}} =\{\boldsymbol{c}, \boldsymbol{h}, \boldsymbol{\underline{\sigma}}, \boldsymbol{\overline{\sigma}}\}$ with  $\boldsymbol{h} \in \mathbb{R}^{P \times M}$, $\boldsymbol{\underline{\sigma}} \in \mathbb{R}^{P \times M}$, $\boldsymbol{\overline{\sigma}} \in \mathbb{R}^{P \times M}$.
\end{itemize}

During the training of IT2-FLS, $\boldsymbol{\theta}_{\boldsymbol{A}} \in \boldsymbol{C}$ must satisfy constraints as they are defining of IT2-FSs \cite{beke}. These constraints are $h_{p, m} \in[0,1],  \underline{\sigma}_{p, m} \leq \bar{\sigma}_{p, m}$ for $\forall p, m$. Yet, as built-in DL optimizers are defined for unconstrained problems, we deploy the following tricks to define an unconstrained training problem: 
\begin{itemize}
    \item $h_{p, m} = \operatorname{sig}({h'_{p, m}})$, where $\operatorname{sig}(\cdot)$ is the sigmoid function.
    \item $\underline{\sigma}_{p, m}=\sigma_{p, m}^{\prime}-\left|\Delta\right|\text{ , } \overline{\sigma}_{p, m}=\sigma_{p, m}^{\prime}+\left|\Delta\right|$
\end{itemize}
 where $\left\{h_{p, m}^{\prime}, \sigma_{p, m}^{\prime}, \Delta_{\mathrm{A}}\right\} \in [-\infty, \infty]$. Now, \eqref{loss_cons} can be defined within widely used DL frameworks such as Matlab and PyTorch. 

\subsection{Scaling T1-FLS Inference to Mini-batches}

Here, we present an efficient T1-FLS inference implementation for a mini-batch $S^{(i)}$. For a given $\boldsymbol{\theta}$, we first define $\{c^{Rep},\sigma^{Rep}\} \in \mathbb{R}^{P\times M \times B}$ and $a_0^{Rep} \in \mathbb{R}^{P\times B}$: 
\begin{equation}
   \{c^{Rep},\sigma^{Rep}\} = \{c,\sigma\} \otimes \mathbb{I}^{1\times 1 \times B}\text{ , }
    a_0^{Rep} = a_0 \otimes \mathbb{I}^{1 \times B}
\end{equation}
and then the inference is accomplished with the following steps:
\begin{itemize}
    \item Calculate $\mu_{A} \in \mathbb{R}^{P\times M \times B}$ for a $S^{(i)}$ with:
        \begin{equation}
            \mu_{A} = \exp\left((x'^{Rep} \ominus c^{Rep})^{2} \oslash 2(\sigma^{Rep})^{2}\right)
        \end{equation}
        by computing $x'^{Rep} \in \mathbb{R}^{P\times M \times B}$ from:
        \begin{equation}
            x'^{Rep} =  x' \otimes \mathbb{I}^{P\times 1  \times 1} 
        \end{equation}
        with $x' \in \mathbb{R}^{1\times M\times B}$
        \begin{equation}
           x' = Permute(x^{(i)}_{1:B}) \text{ from } (0,1,2) \text{ to } (2,0,1)  
        \end{equation}
   \item Calculate $f_p, \forall p,m,n$: 
        \begin{equation}
            {f} = \prod_{m=1}^{M}{\mu_{A}}\text{ , }{f} \in \mathbb{R}^{P\times 1\times B}
        \end{equation}
    \item The output $y_{T1} \in \mathbb{R}^{1\times 1 \times B}$ for a $S^{(i)}$ is calculated via:
        \begin{equation}
            y_{T1} = \sum\nolimits_{P}(f_{norm} \odot y'_{p})
        \end{equation}
        where ${f_{norm}} \in \mathbb{R}^{P\times 1\times B}$ is obtained through:
       \begin{gather}             
        f_{norm} = f  \oslash f_{sum}^{Rep}  \notag
         \\  
         f^{Rep}_{sum} = f_{sum} \otimes \mathbb{I}^{P\times 1 \times 1} \text{ , }{f_{sum}^{Rep}} \in \mathbb{R}^{P\times 1\times B}
         \\
        f_{sum} = \sum\nolimits_{P}{f}\text{ , }{f_{sum}} \in \mathbb{R}^{1\times 1\times B}  \notag
       \end{gather}
        and $ y'_{p} \in \mathbb{R}^{P\times 1\times B}$ derived by first computing             \begin{equation}\label{eq:consequent}
            y_{p} = a\boldsymbol{x}_{1:B}^{(i)} \oplus a_0^{Rep}\text{ , } y_{p} \in \mathbb{R}^{P\times B}
        \end{equation}
        followed by the permutation:      \begin{equation}\label{eq:consequent_permuted}
            y'_{p}= Permute(y_{p}) \text{ from } (0,1,2) \text{ to } (0,2,1).
        \end{equation}        
\end{itemize}

\subsection{Scaling IT2-FLS Inference to Mini-batches}

The inference time and complexity of the IT2-FLS is high due to the iterative nature of KMA which also requires a sorting procedure. It has been shown in \cite{KUMBASAR20171} that \eqref{IT2_lower_upper} can be reformulated via ${u_p} \in {0,1}$ which defines an equivalent $f_{p} = \overline{f}_{p} u_p + \underline{f}_{p} (1-u_p)$. Yet, this new formulation creates $P$ new optimization variables and still requires finding $\underline{y}_{IT2}$ and $\overline{y}_{IT2}$ iteratively. 

In this paper, we present an efficient IT2-FLS inference implementation by eliminating the required optimization problems in \eqref{IT2_lower_upper} by evaluating
\begin{gather} \label{XZ}
Y(\boldsymbol{u})=X(\boldsymbol{u}) \oslash Z(\boldsymbol{u})
\\
\begin{aligned}
    X(\boldsymbol{u}) =& \alpha_{0}+\alpha \boldsymbol{u}: \quad \alpha_{0} = \sum\nolimits_{p}^{P} {y}_{p} \underline{f}_{p}\text{ ; }\alpha_{p} = {y}_{p} (\overline{f}_{p} - \underline{f}_{p})\text{ , } p=1,...,P
\\
    Z(\boldsymbol{u}) =& \beta_{0}+ \beta\boldsymbol{u}: \quad \beta_{0} = \sum\nolimits_{p}^{P} \underline{f}_{p}\text{ ; }\beta_{p} = (\overline{f}_{p} - \underline{f}_{p})\text{ , } p=1,...,P 
\end{aligned}
\end{gather}

\noindent with $\boldsymbol{u} \in \mathbb{R}^{P\times 2^P}$ that defines all binary combinations $u_p$ as follows:
\begin{equation}\label{u_rep}
    \boldsymbol{u} = \prod_{p=1}^{P} \{0,1\}.
\end{equation}
Here, $\prod$ denotes the cartesian product. For instance, $\boldsymbol{u}$ for $P = 3$ is as follows:
\begin{equation}
    \boldsymbol{u} = \prod_{p=1}^{3} \{0,1\} = 
    \begin{bmatrix}
    0 & 0 & 0 & 0 & 1 & 1 & 1 & 1\\
    0 & 0 & 1 & 1 & 0 & 0 & 1 & 1\\
    0 & 1 & 0 & 1 & 0 & 1 & 0 & 1
    \end{bmatrix}	
\end{equation}
As $Y(\boldsymbol{u})$ includes all possible solutions, we can obtain seamlessly $\underline{y}_{IT2}$ and $\overline{y}_{IT2}$: 
\begin{equation}
\underline{y}_{IT2}=\min(Y(\boldsymbol{u}))\text{ , }\overline{y}_{IT2}=\max(Y(\boldsymbol{u}))
\end{equation}
Note that, we chose this inference implementation because GPUs are renowned for their efficiency in executing matrix operations.

Let us now extend the IT2-FLS inference to a mini-batch. For a given $\boldsymbol{\theta}$, we define the variables $\{c^{Rep},h^{Rep}\} \in \mathbb{R}^{P\times M \times B}$ and $\{\underline{\sigma}^{Rep},\overline{\sigma}^{Rep}\} \in \mathbb{R}^{P\times M \times B}$: 
\begin{equation}
   \{c^{Rep},h^{Rep}\}  = \{c,h\} \otimes \mathbb{I}^{1\times 1 \times B}\text{ , }
    \{\underline{\sigma}^{Rep},\overline{\sigma}^{Rep}\} = \{\underline{\sigma},\overline{\sigma}\} \otimes \mathbb{I}^{1\times 1 \times B}
\end{equation} 
and then the inference is accomplished with the following steps: 
\begin{itemize}
    \item Calculate $\overline{\mu}_{\tilde{A}}  \in \mathbb{R}^{P\times M \times B}$ and $\underline{\mu}_{\tilde{A}}  \in \mathbb{R}^{P\times M \times B}$ as follows:
    \begin{equation}
            \begin{split}
            \overline{\mu}_{\tilde{A}} = \exp\left((x'^{Rep} \ominus c^{Rep})^{2} \oslash 2(\overline{\sigma}^{Rep})^{2}\right)\\
            \underline{\mu}_{\tilde{A}} = h^{Rep} \otimes \exp\left((x'^{Rep} \ominus c^{Rep})^{2} \oslash 2(\underline{\sigma}^{Rep})^{2}\right)
            \end{split}
    \end{equation}
        by performing the following operations on the input: 
        \begin{gather}
            x' = Permute(x^{(i)}_{1:B}) \text{ from } (0,1,2) \text{ to } (2,0,1)\text{ , } x' \in \mathbb{R}^{1\times M\times B} \\
            x'^{Rep} = x' \otimes \mathbb{I}^{P\times 1 \times 1}\text{ , } x'^{Rep} \in \mathbb{R}^{P\times M \times B}
            \end{gather}
    \item Calculate $\{\underline{f}, \overline{f}\}  \in \mathbb{R}^{P\times 1\times B}$ for the mini-batch: 
        \begin{equation}
            \underline{f} = \prod_{m=1}^{M}{\underline{\mu}_{\tilde{A}}} \text{ , }\overline{f} = \prod_{m=1}^{M}{\overline{\mu}_{\tilde{A}}}
        \end{equation}  
     \item Compute the component $\boldsymbol{Z}(u) \in \mathbb{R}^{1\times 2^P \times B}$  of \eqref{XZ}: 
          \begin{equation}
            \boldsymbol{Z}(u) = \boldsymbol{\beta}_{0}^{Rep} \oplus \boldsymbol{\beta}'
        \end{equation} 
        by evaluating $\boldsymbol{\beta}_{0}^{Rep} \in \mathbb{R}^{1\times 2^P \times B}$ via:
       \begin{equation}
            \boldsymbol{\beta}_{0}^{Rep} = \boldsymbol{\beta}_{0} \otimes \mathbb{I}^{1\times 2^P \times 1}\text{ , }\boldsymbol{\beta}_{0} = \sum_{p=1}^{P}\underline{f}\text{ , }\boldsymbol{\beta}_{0} \in \mathbb{R}^{1\times 1 \times B}
        \end{equation}
        and $\boldsymbol{\beta}' \in \mathbb{R}^{1\times 2^P \times B}$ via:
        \begin{gather}
            \boldsymbol{\beta}' = Permute(\boldsymbol{\beta}) \text{ from } (0,1,2) \text{ to } (2,1,0)\text{ , }
            \boldsymbol{\beta} = \Delta f' u \text{ , } \boldsymbol{\beta} \in \mathbb{R}^{B\times 2^P}
        \end{gather}
        where $\Delta f' \in \mathbb{R}^{B\times P}$ is derived from 
        \begin{equation}
            \Delta f = \overline{f} \ominus \underline{f}\text{ , }\Delta f \in \mathbb{R}^{P\times 1\times B}
        \end{equation}
        through the following permutation
        \begin{equation}
            \Delta f' = Permute(\Delta f) \text{ from } (0,1,2) \text{ to } (2,0,1).
        \end{equation}

    \item Calculate the component $\boldsymbol{X}(u) \in \mathbb{R}^{1\times 2^P \times B}$  of \eqref{XZ}:   
        \begin{equation}
            \boldsymbol{X}(u) = \boldsymbol{\alpha}_{0}^{Rep} \oplus \boldsymbol{\alpha}'.
        \end{equation} 
        Here, $\boldsymbol{\alpha}' \in \mathbb{R}^{1\times 2^P \times B}$ is obtained from $\boldsymbol{\alpha}^{Tp} \in \mathbb{R}^{B\times P}$ as follows:
        \begin{gather}
             \boldsymbol{\alpha}' = Permute (  \boldsymbol{\alpha}) \text{ from } (0,1,2) \text{ to } (2,1,0)          \\
                         \boldsymbol{\alpha} = \boldsymbol{\alpha}^{Tp}u
            \text{ , }\boldsymbol{\alpha} \in \mathbb{R}^{B\times 2^P} \text{ , }
            \boldsymbol{\alpha}^{Tp} = (y''_{p} \odot \Delta f')
        \end{gather} 
        where $y''_{p} \in \mathbb{R}^{B\times P}$ is obtained from $y_{p}$ (given in \eqref{eq:consequent}) by performing:
        \begin{equation}
            y''_{p} = Permute ( y_{p} ) \text{ from } (0,1,2) \text{ to } (2,0,1)
        \end{equation}
        and $\boldsymbol{\alpha}_{0}^{Rep} \in \mathbb{R}^{1\times 2^P \times B}$ is computed using $y'_{p}$ (given in \eqref{eq:consequent_permuted}) through: 
        \begin{gather}
            \boldsymbol{\alpha}_{0}^{Rep} = \boldsymbol{\alpha}_{0} \otimes \mathbb{I}^{1\times 2^P \times 1} \text{ , }
            \boldsymbol{\alpha}_{0} = \sum\nolimits_{P}(y'_{p} \odot \underline{f})\text{ , }\boldsymbol{\alpha}_{0} \in \mathbb{R}^{1\times 1 \times B}         
        \end{gather}
    \item Compute $\boldsymbol{Y}(u) \in \mathbb{R}^{1\times 2^P \times B}$
        \begin{equation}
             \boldsymbol{Y}(u) = \boldsymbol{X}(u) \oslash \boldsymbol{Z}(u)
        \end{equation}
    and obtain $\underline{y}_{IT2}\in \mathbb{R}^{1\times 1 \times B}$ and $\overline{y}_{IT2}\in \mathbb{R}^{1\times 1 \times B}$ simply with
    \begin{equation}
            \underline{y}_{IT2} = \min(\boldsymbol{Y}(u)(:,:,b)) \text{ , }
            \overline{y}_{IT2} = \max(\boldsymbol{Y}(u)(:,:,b)) \text{ , } \forall b
    \end{equation}
    to calculate $y_{IT2} \in \mathbb{R}^{1\times 1 \times B}$:
        \begin{equation}
            y_{IT2} = (\underline{y}_{IT2} \oplus \overline{y}_{IT2})/2
        \end{equation}     
\end{itemize}

\noindent \textbf{Remark:} Note that while the FLS inference is demonstrated for a single output, the implementation can be extended to accommodate multi-output FLSs.

\section{Performance Analysis}
We analyze the FLS learning performances on the Power Plant (CCPP)$(M=4, D=1, N=9568)$, Boston Housing (BH)$(M=13, D=1, N=506)$, and Energy Efficiency (ENB)$(M=8, D=2, N=768)$ datasets. All datasets are preprocessed by Z-score normalization, and we partition them into a 70\% training set and a 30\% test set. FLSs are trained for 100 epochs using MATLAB\textsuperscript{\textregistered} Deep Learning Toolbox \footnote{The MATLAB implementation. [Online]. Available:\\ \href{https://github.com/atakoklu/Efficient-Learning-of-Fuzzy-Logic-Systems-for-Large-Scale-Data-Using-Deep-Learning}{https://github.com/atakoklu/Efficient-Learning-of-Fuzzy-Logic-Systems-for-Large-Scale-Data-Using-Deep-Learning}}.  All tests were conducted on a PC equipped with an Intel i9-7920 CPU with 12 physical cores and an NVIDIA GTX 1080 Ti GPU.

Performance evaluation includes training times and testing RMSE values. As tabulated in Table \ref{tab:testler}, our implementations of T1-FLS and IT2-FLS (abbreviated as IT2-$f$KM) have accomplished very short training times. Improvement in the training time is visible on $f$KM compared to the KMA. Our implementation of IT2-FLS was 7218 times faster than the KMA since our implementation computed all of the possible combinations of the KMA in parallel in GPU. Moreover, our implementation is more robust across varying numbers of $P$ and $D$.
\begin{table}[htbp]    
    \centering
        \caption{Perfomance Anaylsis}
        \begin{center}
        \begin{tabular}
        {|c|c||c|c|c||c|c|c|c|c|c|}
        \hline
        &&\multicolumn{3}{c||}{Training Time}&\multicolumn{6}{c|}{Test RMSE}\\
        \hline
        \hline
        &$P$&T1&IT2-KM&IT2-$f$KM&\multicolumn{2}{c|}{T1}&\multicolumn{2}{c|}{IT2-KM}&\multicolumn{2}{c|}{IT2-$f$KM}\\
        \hline
        \hline
        \multirow{3}{*}{\begin{sideways}CCPP\end{sideways}}&5&11s&18h 16m&18s&\multicolumn{2}{c|}{0.246}&\multicolumn{2}{c|}{0.242}&\multicolumn{2}{c|}{0.242}\\
        &10&11s&38h 6m&19s&\multicolumn{2}{c|}{0.241}&\multicolumn{2}{c|}{0.237}&\multicolumn{2}{c|}{0.237}\\
        &15&11s&57h 11m&50s&\multicolumn{2}{c|}{0.238}&\multicolumn{2}{c|}{0.235}&\multicolumn{2}{c|}{0.235}\\
        \hline
        \hline
        \multirow{3}{*}{\begin{sideways}BH\end{sideways}}&5&5s&50m&8s&\multicolumn{2}{c|}{0.450}&\multicolumn{2}{c|}{0.442}&\multicolumn{2}{c|}{0.442}\\
        &10&5s&1h 38m&8s&\multicolumn{2}{c|}{0.426}&\multicolumn{2}{c|}{0.428}&\multicolumn{2}{c|}{0.428}\\
        &15&5s&2h 28m&9s&\multicolumn{2}{c|}{0.416}&\multicolumn{2}{c|}{0.420}&\multicolumn{2}{c|}{0.420}\\
        \hline
        \hline
        \multirow{3}{*}{\begin{sideways}ENB*\end{sideways}}&5&7s&3h 2m& 12s&0.136&0.203&0.082&0.169&0.082&0.169\\
        &10&7s& 6h 50m& 13s&0.102&0.173&0.130&0.175&0.130&0.175\\
        &15&7s & 9h 57m& 45s&0.068&0.153&0.149&0.204&0.149&0.204\\
        \hline
        \multicolumn{11}{@{}l}{*The RMSE values for output-1 and output-2 are provided.}
        \end{tabular}
    \end{center}
\label{tab:testler}
\end{table}  
\vspace{-30pt}
\section{Conclusion and Future Work}
This paper presented computationally efficient inference implementations for FLS inferences to train FLSs within DL frameworks. Thanks to our efficient implementations, we were able to seamlessly solve the learning problem of the FLSs by leveraging automatic differentiation and DL optimizers. 

We have presented all the details on how to develop a DL framework for FLSs. We first provided parameterization tricks for the LPs to transform the constraint learning problem of FLSs into unconstrained ones so that training via unconstrained DL optimizers is feasible. Then, we focused on the inference computations and presented efficient mini-batched inferences for FLSs. For IT2-FLSs, we eliminated the iterative nature of KMA by developing an inference implementation that efficiently computes all possible combinations of the type-reduced set by harnessing the power of parallelization. The efficacy of the FLS learning framework is showcased through its application on various benchmark datasets commonly used for learning tasks. The comparative results illustrated that the proposed implementations significantly improved training time without compromising the accuracy performance of the FLSs. 

In future work, we aim to extend this implementation to the PyTorch framework and integrate it into the TinyML framework.


\end{document}